# Validation of Diagnostic Artificial Intelligence Models for Prostate Pathology in a Middle Eastern Cohort

Peshawa J. Muhammad Ali[1,2*], Navin Vincent[3*], Saman S. Abdulla[4,5], Han N. Mohammed Fadhl[6], Anders Blilie[7,8], Kelvin Szolnoky[9], Julia Anna Mielcarz[3], Xiaoyi Ji[9], Nita Mulliqi[3], Abdulbasit K. Al-Talabani[1‡], Kimmo Kartasalo[3‡]

1. Department of Software Engineering, Faculty of Engineering, Koya University, Koya 44023, Kurdistan Region - F.R. Iraq
2. Department of Mechanical and Manufacturing Engineering, Faculty of Engineering, Koya University, Koya 44023, Kurdistan Region - F.R. Iraq
3. Department of Medical Epidemiology and Biostatistics, SciLifeLab, Karolinska Institutet, Stockholm, Sweden
4. College of Dentistry, Hawler Medical University, Erbil, Kurdistan Region, Iraq
5. PAR Private Hospital, Erbil, Kurdistan Region, Iraq
6. College of Dentistry, University of Sulaimani, Sulaymaniyah, Kurdistan Region, Iraq
7. Department of Pathology, Stavanger University Hospital, Stavanger, Norway
8. Faculty of Health Sciences, University of Stavanger, Stavanger, Norway
9. Department of Medical Epidemiology and Biostatistics, Karolinska Institutet, Stockholm, Sweden

\* These authors contributed equally to this work.
‡ These authors jointly supervised the work.

Corresponding author: Kimmo Kartasalo, kimmo.kartasalo@ki.se.

# Abstract

**Background**

Artificial intelligence (AI) is improving the efficiency and accuracy of cancer diagnostics. The performance of pathology AI systems has been almost exclusively evaluated on European and US cohorts from large centers. For global AI adoption in pathology, validation studies on currently under-represented populations – where the potential gains from AI support may also be greatest – are needed. We present the first study with an external validation cohort from the Middle East, focusing on AI-based diagnosis and Gleason grading of prostate cancer.

**Methods**

We collected and digitised 339 prostate biopsy specimens from the Kurdistan region, Iraq, representing a consecutive series of 185 patients spanning the period 2013-2024. We evaluated a task-specific end-to-end AI model and two foundation models in terms of their concordance with pathologists and consistency across samples digitised on three scanner models (Hamamatsu, Leica, and Grundium).

**Findings**

Grading concordance between AI and pathologists was similar to pathologist-pathologist concordance with Cohen's quadratically weighted kappa 0.801 vs. 0.799 (p=0.9824). Cross-scanner concordance was high (quadratically weighted kappa > 0.90) for all AI models and scanner pairs, including low-cost compact scanner.

**Interpretation**

AI models demonstrated pathologist-level performance in prostate histopathology assessment. Compact scanners can provide a route for validation studies in non-digitalised settings and enable cost-effective adoption of AI in laboratories with limited sample volumes. This first openly available digital pathology dataset from the Middle East supports further research into globally equitable AI pathology.

**Funding**

SciLifeLab & Wallenberg Data Driven Life Science Program, Instrumentarium Science Foundation, Karolinska Institutet Research Foundation.

# Research in context

**Evidence before this study**

We found 119 and 34 records searching PubMed Central and PubMed, respectively, on December 8, 2025, for peer-reviewed articles in English using the search terms "prostate cancer"[Title/Abstract] AND ("histopath*" OR "histol*" OR "whole slide image" OR "WSI") AND ("artificial intelligence"[Title/Abstract] OR "machine learning"[Title/Abstract] OR "deep learning"[Title/Abstract] OR "AI"[Title/Abstract]), combined with the terms "Middle East", "UAE" and a list of 16 countries in the region. After full-text review, we identified one study with data from the region (Iran). The study relied on a private dataset of photomicrographs rather than whole slide images, lacked clear scanning parameters and patient counts, and was described by the authors as unsuitable for standard benchmarking. We also examined the reference lists of the included papers and websites of prostate pathology software vendors, and found one article with development and internal validation on Israeli data. We concurrently searched five major Arabic-language academic databases (Iraqi Academic Scientific Journal, E-Marefa, Al-Maqsurah, ALMANDUMA, and the Iraqi Digital Repository) for the term "prostate cancer" (سرطان البروستات). These yielded 27, 51, 2, 38, and 69 articles, respectively, but a sub-search for AI-related terminology (الذكاء الأصطناعي) revealed zero relevant hits in four of the databases. The Iraqi Digital Repository returned one computational study outside histopathology. We found no public datasets from the Middle East suited for development or validation of AI for prostate histopathology.

**Added value of this study**

To our knowledge, this is the first study to externally validate the performance of AI systems, including both task-specific and foundation models, on a prostate cancer cohort from the Middle East, a region historically excluded from computational pathology research. We demonstrated that AI models trained exclusively on Scandinavian data generalised effectively to Iraqi patients, achieving diagnostic and grading accuracy comparable to experienced urological pathologists. We established that this performance is maintained across different digitisation hardware, including a low-cost compact scanner suitable for resource-limited settings. Furthermore, to support further research into globally equitable AI, we have released the first open-access digital pathology dataset from this region.

**Implications of all the available evidence**

Our findings challenge the assumption that AI models are inevitably biased against populations absent from their training data, suggesting that high-quality prostate histopathology AI is robust to demographic shifts. This has implications for global health equity: it indicates that advanced diagnostic tools developed in high-income countries may be validated and safely deployed in emerging economies without the need for massive local retraining efforts. Moreover, by validating these tools on portable, low-throughput scanners, we outline a pragmatic, cost-effective pathway for non-digital laboratories to adopt AI.

# Introduction

Artificial intelligence (AI) is increasingly adopted in medicine, and anatomical pathology is among the fields most amenable to AI-based approaches.[1] Successful examples include cancer diagnosis, prediction of molecular biomarkers, and prognostication of outcomes. Recently, computational pathology has progressed rapidly with the emergence of foundation models (FM), i.e. AI models applicable across different tasks and tissues with minimal refinements, in contrast to the conventional approach of tailor-made models for each problem. Clinically, AI is expected to improve patient safety, reduce diagnostic inter-observer variability, and support health care struggling with pathology workforce shortages.

Medical devices often work best for the people they were built for. The exclusion of under-represented patient groups contributes to 'health data poverty' and can reduce the accuracy of AI-enabled devices.[2] In computational pathology, AI has almost exclusively been developed on North American and European data. Failures to generalise across patient demographics,[3] laboratories, or even digital pathology scanner models[4] are widely reported. Establishing the safe and accurate operation of AI-based tools by validation studies before clinical application in new settings is therefore crucial, but limited research outside Western settings excludes many of the patients who could benefit the most from this technology.

Here, we focused on one of the most widely studied problems in computational pathology: diagnosis and grading of prostate cancer in whole slide images (WSI).[5] Gleason grading of cancer is crucial in therapeutic decision-making, but is plagued by inter-observer variability. Several studies have demonstrated AI can perform this task consistently[6–8] and commercial software is already available (e.g., Paige Prostate,[9–11] DeepBio Dx,[12] Halo Prostate,[13] INIFY Prostate,[14] and Ibex Galen Prostate[15–17]). These models have been validated mainly in European and North American populations or other developed, digitalised settings **(Supplementary Table 1)**. Our objective was to evaluate whether AI systems developed using datasets from Scandinavia maintain their diagnostic performance in a Middle Eastern population previously not represented in computational pathology research – the Kurdistan Region of Iraq, where digital pathology is not yet available and evidence supporting AI pathology is lacking.

The global burden of prostate cancer is rising: annual new cases are projected to nearly double by 2050 under constant age-specific rates.[18] Marked outcome disparities exist when contrasting high-income, digitally mature settings with under-resourced environments: for example mortality to incidence ratios are 0.13 vs 0.39 in Sweden and Iraq, respectively **(Supplementary Table 2)**.[19] This is consistent with Iraq being classified among fragile states in comparative analyses of cancer control capacity.[20] Pathology services remain under-resourced and under-digitised: diagnosis still relies mainly on glass slides, access to scanners and reliable IT is uneven, AI use in histopathology is uncommon, local datasets are scarce, and research funding is limited. Successful adoption of AI could potentially alleviate some of these issues and improve pathology practice in the region, but models built elsewhere may be miscalibrated on Iraqi samples due to differences in patient demographics and sample preparation. Previously reported geographical variations in prostate cancer characteristics support this hypothesis.[21–23] Against the backdrop of rising global burden, it is ethically urgent to validate imported AI in under-represented settings such as Iraq to avoid widening inequities.

Even if AI models generalise to currently under-represented populations, a major hurdle remains: AI requires a digitised laboratory. Implementing full digitisation demands substantial time and investments into high-throughput scanners, image management systems, high-speed networks, and workstations and training for pathologists. Most laboratories cannot justify these investments without prior evidence of AI benefit. Ironically, the absence of digitisation prevents generating the evidence that would be needed to justify digitisation. This catch-22 effectively excludes the very settings that stand to benefit most from AI in pathology. As a means of escaping this conundrum, we suggest low-cost compact scanners as a vehicle for conducting validation studies in currently non-digitalised laboratories.

In this study, we collected and digitised a consecutive series of prostate core needle biopsies from Erbil, Kurdistan Region of Iraq, representing a decade of routine diagnostic practice in the region. Relying on reference grading by three pathologists, we retrospectively evaluated the performance of three AI models on this cohort: a task-specific model (TS)[24] developed at Karolinska Institutet, Sweden, and publicly available FMs UNI[25] and Virchow2[26] pre-trained on North American data. All models were trained to perform Gleason grading using 51,247 WSIs of prostate core needle biopsies from Sweden and Norway, and previously evaluated on 25,601 WSIs from 15 clinical sites in Europe and Australia.[24] In addition, we evaluated the models' cross-scanner generalisation. We hypothesised that (i) the AI systems would perform comparably to pathologists on the Iraqi cohort, and (ii) their predictions would remain consistent across compact and high-throughput scanners. The AI development and validation followed a pre-registered study protocol[27] and the STARD-AI checklist **(Supplementary Table 9)**.[28] Moreover, we publicly share the dataset of 1,017 WSIs to provide the computational pathology community with the first open dataset from the Middle East.[29]

# Methods

## Study design and participants

Potentially eligible participants were identified through the pathology archive of PAR Private Hospital, Erbil, Kurdistan Region of Iraq. All pathology reports mentioning prostate cancer were retrieved and reviewed, followed by restriction to suspected adenocarcinoma and further to core needle biopsies by excluding prostatectomies and transurethral resections of the prostate. The dataset represents a consecutive series of all eligible prostate biopsy cases diagnosed between mid-2013 and mid-2024, included without any sampling, and mirrors the intended-use population in this region (**Fig. 1**).

The dataset comprises 339 glass slides representing 185 patients. Each slide contains an average of six biopsy cores, extracted with transrectal ultrasound-guided biopsy, and stained with haematoxylin & eosin. The minimum, mean, and maximum ages at biopsy were 33, 68, and 95 years, respectively, and the interquartile range was 64-74 years (**Table 1**). Two glass slides were available for 154 patients (representing cores from the right and left side of the prostate), while 31 patients had all cores on one slide. Stain fading was observed in older slides, and to produce samples reflecting those that would be analysed in a prospective setting, fresh slides were prepared from the original paraffin blocks by re-cutting (and re-embedding where necessary). De-identified samples were shipped to Karolinska Institutet for scanning.

The sample collection was approved by the Ethical Committee of PAR Hospital (permit 1002/07072024). Scanning of the samples at Karolinska Institutet was approved by the Swedish Ethical Review Authority (permit 2019-05220). Informed consent was waived by the ethical review board due to the use of de-identified specimens in a retrospective setting.

## Slide scanning

The slides were scanned at Karolinska Institutet using three scanners: Hamamatsu NanoZoomer 2.0 HT, Leica Aperio GT450 DX, and the compact Grundium Ocus40. Manual rescans were carried out when automated focusing failed. Slides were scanned at 40× magnification, yielding resolutions of 0.22 µm/pixel (Hamamatsu), 0.26 µm/pixel (Aperio), and 0.25 µm/pixel (Grundium). The WSIs were saved in .ndpi (Hamamatsu) and .svs (Leica and Grundium) formats.

## Reference standard

Three pathologists assessed the slides. Pathologist I (S.A.) is a general pathologist with 22 years of experience and a special interest in prostatic pathology. He serves as lecturer at Hawler Medical

University, as consultant pathologist at PAR Hospital, and as senior examiner for the Royal College of Pathologists in the UK. Pathologist II (H.M.) is a general pathologist with 15 years of experience and is currently affiliated with the University of Sulaimani in Sulaymaniyah, Iraq. Pathologist III (A.B.) is a uropathologist with 6 years of experience at Stavanger University Hospital in Stavanger, Norway.

All pathologists conducted digital assessments of the Hamamatsu scans using Cytomine[30] and assigned slide-level Gleason scores (GS) (using 0+0 for benign cases) **(Table 1)**. The GS were additionally converted to International Society for Urological Pathology (ISUP) grades. S.A. graded all 339 slides, H.M. graded 337 slides, and A.B. evaluated a random subset of 59 slides stratified according to the ISUP grades assigned by S.A. The pathologists were blinded to the patients' clinical characteristics, each other's assessments, and AI outputs. Alternative diagnoses were reported during the original diagnostic assessment **(Supplementary Table 6-7)**.

Following AI analysis, slides where any AI models showed marked errors compared to both S.A. and H.M. were further examined. We followed a pre-specified criterion[27] for the most clinically significant errors: benign samples AI-predicted as ISUP grade ≥2 and samples of ISUP grade ≥2 AI-predicted as benign **(Supplementary Table 8)**. Slides where at least two out of three AI models exhibited such an error were re-assessed and described by A.B., blinded to the other assessments and the AI models.

## Artificial intelligence model

The AI models were implemented in PyTorch as described earlier[24] and no significant custom code was developed for this study. In brief, tissue was detected using a segmentation model based on a UNet++ architecture. Patches of 256×256 px extracted at 1.0 µm/px from tissue regions were used as input for the diagnostic models. All models used an attention-based multiple instance learning (ABMIL) framework relying on slide-level labels for weak supervision. In the TS model, patch-level features were extracted with a trainable encoder and attention-aggregated into slide-level representations. Downstream classification layers were used to predict the primary and secondary Gleason patterns, further translated into GS and ISUP grades. The same design was applied to the UNI[25] and Virchow2[26] FMs keeping their encoders frozen and training the ABMIL and classification layers identically to the TS setup. All AI inferences were performed solely on WSIs without access to clinical information or reference standards. For details, see **Supplementary Methods**.

## Statistical analysis

The AI development and validation followed a protocol-based study design, where information leakage between development and validation data was eliminated and all primary analyses were pre-specified.[27] Here, we followed this design by handling the Iraqi cohort identically to earlier external validation cohorts ensuring full independence from development data in terms of: i) Geographic location, ii) Patients, iii) Laboratory, iv) Scanner. Sample size was dictated by the inclusion of all available eligible archival cases rather than based on formal power calculations. Statistical calculations were performed in Python using NumPy v1.24.0, scikit-learn v1.2.2, and Pandas v1.5.3.

To assess AI agreement with reference standards in cancer detection, we reported sensitivity (true positive rate), specificity (true negative rate), and area under the receiver operating characteristic (ROC) curve (AUC). Concordance in grading was measured primarily with Cohen's quadratically weighted kappa (QWK) and secondarily with linearly weighted kappa (LWK). 95% confidence intervals were derived by bootstrapping over 1000 replicates on a per-observer basis. No indeterminate or AI-ungradable slides were encountered and all cases yielded valid predictions.

To evaluate whether the mean concordance of AI systems with pathologists differed significantly from that between pairs of pathologists, we conducted bootstrap hypothesis testing. For each observer pair, we resampled from the original paired grades to construct bootstrap distributions of Cohen's kappa (both QWK and LWK for ISUP and GS), then pooled these distributions separately for pathologist-pathologist pairs and AI-pathologist pairs. We used a balanced sampling strategy targeting ~30,000 total samples (≈15,000 per group): three human–human pairs × 5,000 samples/pair = 15,000, and nine AI–human pairs × 1,666 samples/pair ≈ 14,994. The null hypothesis (H0) was that the mean kappa for AI–human pairs equals the mean kappa for human–human pairs. We derived a two-tailed bootstrap p-value from the empirical distribution of mean differences (AI−human). We also reported Cohen's d as an effect size, computed as the observed mean difference in kappa divided by the pooled bootstrap standard deviation of the two groups. Statistical significance was assessed at $\alpha = 0.05$.

Role of the funding source

# Results

## Diagnostic AI developed on European data generalises to a Middle Eastern setting

We first compared the AI models' cancer diagnoses to pathologist I. The WSIs acquired with the Hamamatsu scanner were used for these primary analyses. All models demonstrated strong performance in discriminating between benign and malignant samples with AUC from 0.916 (UNI) to 0.966 (TS) **(Figure 2)**. In terms of binary benign vs. malignant predictions, this translates to sensitivities from 97.1% (TS) to 98.9% (Virchow2) for detecting malignancy **(Table 2)**. The FMs produced more false positive diagnoses with specificities of 85.4% (UNI) and 88.4% (Virchow2) compared to 94.5% for the TS model. Repeated analysis against pathologist II yielded comparable results with somewhat lower diagnostic agreement (**Supplementary Figure 1**, **Supplementary Table 3)**.

In grading, the models achieved QWK values quantifying concordance with pathologist I from 0.788 (UNI) to 0.869 (TS) for ISUP grades, and from 0.669 (UNI) to 0.789 (TS) for GS **(Table 2)**. As is common both for pathologists and AI models, lower QWK were observed when the analysis was restricted to malignant samples only, reflecting the inherent difficulty in grading compared to the binary cancer detection task. Confusion matrices of AI-predicted and pathologist-reported GS are presented in **Figure 2**. Repeated analysis against pathologist II produced comparable results (**Supplementary Figure 1, Supplementary Table 3**). Grading concordance measured in terms of LWK relative to pathologists I and II is reported in **Supplementary Table 4** and **Supplementary Table 5,** respectively. Confusion matrices for ISUP grades against both pathologists are presented in **Supplementary Figure 2**.

We further evaluated the subset of slides associated with marked AI errors compared to the grading by both pathologists I and II. To highlight the errors arguably associated with the most severe consequences for clinical decision making, we adopted a pre-defined definition of significant errors: a benign sample AI-predicted as ISUP grade ≥2 or a sample of ISUP grade ≥2 AI-predicted as benign.[27] In total, the TS, UNI, and Virchow2 models committed 5 (1.5%), 16 (4.7%) and 7 (2.1%) such errors across the 337 slides assessed by both pathologists. When the slides where at least two out of three AI models committed significant errors (n=9 unique slides) were independently re-assessed by pathologist III, the errors were resolved in 4/5 (TS), 8/9 (UNI) and 6/7 (Virchow2) slides when treating his assessment as the reference standard. This inter-observer variation highlights the difficult nature of many of the included cases, as also indicated by the comments from pathologist III **(Supplementary Table 8)**.

## Grading by AI is indistinguishable from grading by pathologists

Having established that AI generalisation to the Middle Eastern cohort is successful, we directly compared AI-pathologist concordance with pathologist-pathologist concordance. This analysis was conducted on the subset of 59 slides assessed by the three pathologists and three AI models, all blinded to each other's grading. We calculated the pairwise Cohen's kappa between all pathologist-pathologist and pathologist-AI pairs, and reported the mean pairwise kappa for each observer. That is, AI-AI concordance is not included in the calculation of the mean pairwise kappa values for AI observers to avoid positive bias in favor of AI models agreeing (potentially erroneously) with each other.

The observer rankings in terms of mean pairwise QWK **(Figure 3)** and LWK **(Supplementary Figure 3)** show that AI models generally performed within the range of inter-pathologist variability. To rigorously evaluate whether the mean AI concordance with pathologists differed systematically from mean pathologist–pathologist concordance, we conducted a formal statistical comparison using bootstrap hypothesis testing. Across all grading schemes and metrics (ISUP linear, ISUP quadratic, GS linear, and GS quadratic), AI systems did not show statistically significant differences in concordance compared with pathologist–pathologist agreement. Concordance for GS was 0.766 for AI versus 0.796 for pathologists (Δ=−0.029; p=0.9822; Cohen's d=−0.426) in terms of QWK, and 0.681 versus 0.657 (Δ=0.025; p=0.9330; Cohen's d=0.398) in terms of LWK. Concordance for ISUP grade was 0.801 for AI versus 0.799 for pathologists (Δ=0.002; p=0.9824; Cohen's d=0.032) in terms of QWK, and 0.723 versus 0.679 (Δ=0.045; two-tailed bootstrap p=0.9908; Cohen's d=0.688) in terms of LWK.

In all cases, the effect sizes ranged from negligible to small–medium (|d| lies between 0.03–0.69), 95% CIs for the bootstrap distributions overlapped substantially (e.g., ISUP–QWK: humans 0.644–0.910 vs AI 0.613–0.903), and all p-values >0.93, indicating that numerical differences lie within expected sampling variability. These findings support that AI-pathologist concordance is statistically indistinguishable from inter-pathologist concordance in this cohort, and occasional AI-human disagreements reflect typical observer variability rather than systematic model error.

### Scanner consistency analysis

To assess the consistency of AI performance across scanner platforms, we evaluated model outputs when the same physical slides were scanned on three devices: the high-throughput Hamamatsu and Leica scanners and the compact Grundium scanner. The point estimates for ISUP and GS cross-scanner concordance in terms of QWK were consistently high for all models, with inter-scanner differences small relative to their 95% CIs **(Figure 4a, 4b)**. The TS model exhibited average QWK values of 0.956 (ISUP) and 0.941 (GS) across scanner pairs, compared to 0.905 (ISUP) and 0.872 (GS) for UNI and 0.929 (ISUP) and 0.899 (GS) for Virchow2. Examination of individual per-slide predictions for ISUP grades **(Figure 4c)** and for GS **(Supplementary Figure 4)** showed that most slides received identical predictions across scanners, and the remaining discrepancies were predominantly one-step grade shifts (e.g., ISUP 2↔3), which are clinically less consequential than discrepancies of several ISUP grades.

## Discussion

This study provides the first evidence that pathology AI models for prostate biopsy assessment, trained on Western datasets, can generalise to a Middle Eastern population thus far not represented in computational pathology. The models' cancer detection performance on the Iraqi cohort was similar to earlier external validation results across 10 European sites,[24] where sensitivities of 90-100% and specificities of 87-97% were observed. While kappa values should not be compared directly between cohorts with varying grade distributions and patient characteristics, the observed AI-pathologist concordance was similar to the European results, where QWK was 0.48-0.90 for GS and 0.62-0.90 for ISUP grades. Discrepancies were primarily between adjacent Gleason/ISUP categories. The few slides with more marked errors were noted by the pathologists to contain various tissue processing artefacts. In all of the re-assessed cases, the third pathologist would have required immunohistochemistry to verify the diagnosis.

Importantly, all three AI models showed concordance within the range of inter-pathologist variability. In the context of the inter-observer analysis, the concordance patterns emphasize that AI disagreement with a single pathologist does not necessarily imply error: observed differences were on par with human-human variability. This observation is particularly relevant for retrospective validations lacking adjudicated consensus diagnoses – the upper bound for measured "accuracy" will always be dictated by both AI capability and the inherent variability of the reference standard. Ultimately, this limitation must be overcome by validating AI against prognostic or treatment-predictive outcomes, rather than solely reproducing subjective human assessments. Despite this limitation, the current study establishes critical evidence for the technical generalisation capacity of prostate pathology AI systems, which is a necessary prerequisite also for the safe application of clinically even more impactful prognostic and predictive models.

In terms of AI methodology, FMs trained in a task-agnostic manner on diverse datasets are frequently positioned as the new standard paradigm for medical AI. Our results are aligned with earlier observations suggesting that while FMs can achieve strong performance in prostate pathology, task-specific AI models tailor-made for this problem can still offer performance comparable or even superior to FMs.[24] In resource-constrained settings, the lower computational demand of simpler task-specific models remains a distinct asset.

One of our key observations is that the evaluated AI models operate reliably across scanner platforms, including a low-cost compact scanner – this perhaps seemingly trivial result is not to be taken for granted in light of well-known challenges with cross-scanner generalisation and is, in fact, crucial for accelerating global adoption of AI pathology. To break the digitisation barrier, standalone compact scanners can be deployed anywhere in minutes without existing digital infrastructure. This allows laboratories to validate AI on their own slides before committing to major investments. In sparsely populated areas, such instruments could also be easily transported to provide world-class AI-supported telepathology service to regions where a centralised digital pathology laboratory is not an economically viable option.

In summary, we showed that AI models for prostate biopsy assessment trained on European data can generalise to a Middle Eastern patient cohort without any tuning or calibration, diagnose cancer comparably to experienced pathologists in this setting, and consistently operate on data from high-throughput and compact scanners. This supports a practical pathway for laboratories in resource-limited settings to evaluate and benefit from computational pathology, advancing diagnostic quality and global equity in access to AI-supported care. This study, and the associated openly available dataset, which is the first of its kind from the Middle East region, hopefully serve as a blueprint for similar validation studies covering other diseases and pathology assessments.

## Contributors
P.J.M.A., S.S.A., A.K.A.-T., and K.K. conceived of the study. P.J.M.A., N.V., and N.M. implemented AI models and conducted the statistical analyses. S.S.A., H.N.M.F., and A.B. conducted pathological assessments. P.J.M.A., N.V., S.S.A., J.A.M., X.J., and N.M. contributed to data collection and data management. K.S. set up and maintained the digital annotation platform. N.M., A.K.A.-T., and K.K. supervised the work. P.J.M.A., N.V., N.M., A.K.A.-T., and K.K. drafted the manuscript. All authors reviewed, edited, and approved the manuscript. P.J.M.A., N.V., N.M., and K.K. had full access to all data.

## Data sharing
The dataset used in this validation study is available under the Creative Commons Attribution 4.0 (CC BY 4.0) license as "Prostate biopsy whole slide image dataset from an underrepresented Middle Eastern population" in the BioImage Archive (BIA): TBA. Images are provided in the vendors' native formats (.svs for Grundium and Leica; .ndpi for Hamamatsu). All AI-predicted outputs are available as Supplementary Data associated with the published article.

## Declaration of interests
N.M. and K.K. are shareholders of Clinsight AB. Other authors declare no competing interests.

## Acknowledgments
The authors acknowledge the collaboration of PAR Private Hospital in Erbil, Iraq, particularly the laboratory team, for providing the prostate cancer samples used in this study. K.K. received funding from the SciLifeLab & Wallenberg Data Driven Life Science Program (KAW 2024.0159), Instrumentarium Science Foundation and Karolinska Institutet Research Foundation. Computations were enabled by the National Academic Infrastructure for Supercomputing in Sweden (NAISS) and the Swedish National Infrastructure for Computing (SNIC) at C3SE partially funded by the Swedish Research Council through grant agreement no. 2022-06725 and no. 2018-05973, and by the supercomputing resource Berzelius provided by the National Supercomputer Centre at Linköping University and the Knut and Alice Wallenberg Foundation.

.

# Figures and Tables

**Table 1: Clinical characteristics and reference-pathologist grading of the study cohort.** Summary of patient- and slide-level characteristics for the 185 patients and 339 prostate biopsy slides included in the study. Gleason scores and ISUP grades represent the assessments of pathologist I, who served as the primary reference standard. Age distribution is shown in 5-year intervals. Values are presented as counts and percentages. ISUP=International Society of Urological Pathology.

| | |
|---|---|
| **Total slides** | 339 |
| **Total patients** | 185 |
| **Gleason score (*n* slides, %)** | |
| Benign | 164 (48.4%) |
| 3+3 | 7 (2.1%) |
| 3+4 | 38 (11.2%) |
| 4+3 | 59 (17.4%) |
| 3+5 | 3 (0.9%) |
| 4+4 | 26 (7.7%) |
| 5+3 | 1 (0.3%) |
| 4+5 | 40 (11.8%) |
| 5+4 | 1 (0.3%) |
| 5+5 | 0 (0.0%) |
| Missing | 0 (0.0%) |
| **ISUP grade (*n* slides, %)** | |
| Benign | 164 (48.4%) |
| ISUP 1 | 7 (2.1%) |
| ISUP 2 | 38 (11.2%) |
| ISUP 3 | 59 (17.4%) |
| ISUP 4 | 30 (8.8%) |
| ISUP 5 | 41 (12.1%) |
| Missing | 0 (0.0%) |
| **Age, years (*n* patients, %)** | |
| <49 | 5 (2.7%) |
| 50–54 | 9 (4.9%) |
| 55–59 | 14 (7.6%) |
| 60–64 | 26 (14.1%) |
| 65–69 | 49 (26.5%) |
| 70–74 | 40 (21.6%) |
| 75–79 | 25 (13.5%) |
| 80–84 | 10 (5.4%) |
| ≥85 | 4 (2.2%) |
| Missing | 3 (1.6%) |

**Table 2: Cancer detection and grading performance metrics of AI models (QWK, primary reference standard).** Specificity, sensitivity, and Cohen's kappa for ISUP grade and Gleason score for three AI systems (task-specific model, UNI foundation model, Virchow2 foundation model). Concordance was computed relative to pathologist I who set the primary reference standard. QWK values are reported for all slides (overall, *n*=339) and for malignant slides (*n*=175) only. Confidence intervals represent 95% bootstrapped estimates. GS=Gleason score, ISUP=International Society of Urological Pathology, QWK=quadratically weighted kappa.

| AI model | Specificity | Sensitivity | QWK - ISUP (Overall) | QWK- ISUP (Malignant) | QWK- GS (Overall) | QWK- GS (Malignant) |
|---|---|---|---|---|---|---|
| Task-specific | 0.945 (0.909-0.977) | 0.971 (0.944-0.994) | 0.869 (0.828-0.905) | 0.640 (0.548-0.722) | 0.789 (0.731-0.842) | 0.588 (0.496-0.675) |
| UNI | 0.854 (0.796-0.905) | 0.983 (0.961-1.000) | 0.788 (0.720-0.852) | 0.663 (0.585-0.735) | 0.669 (0.571-0.758) | 0.599 (0.509-0.680) |
| Virchow2 | 0.884 (0.834-0.931) | 0.989 (0.971-1.000) | 0.848 (0.797-0.893) | 0.619 (0.527-0.693) | 0.753 (0.685-0.817) | 0.525 (0.441-0.608) |

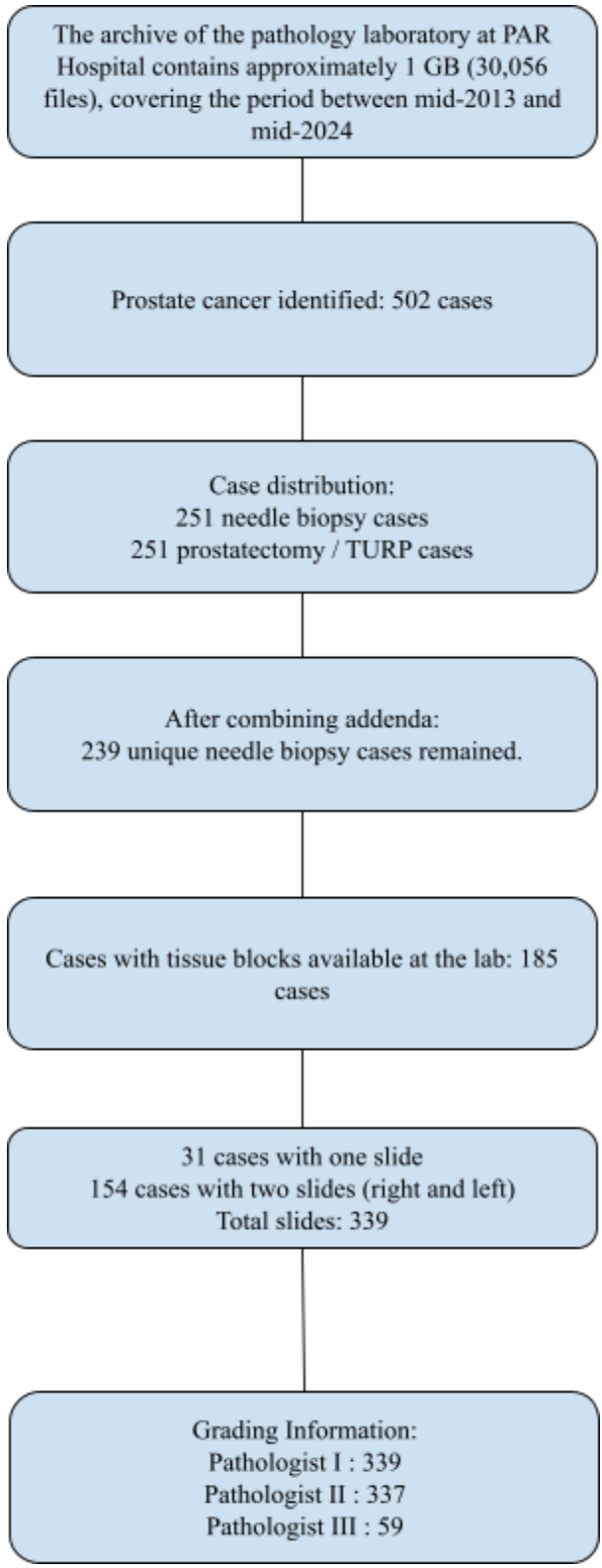

**Figure 1: CONSORT diagram presenting the study cohort.** Flow diagram illustrating the selection of the study cohort from the pathology archive (2013–2024). The final dataset comprises 339 biopsy slides from 185 patients. TURP=transurethral resection of the prostate.

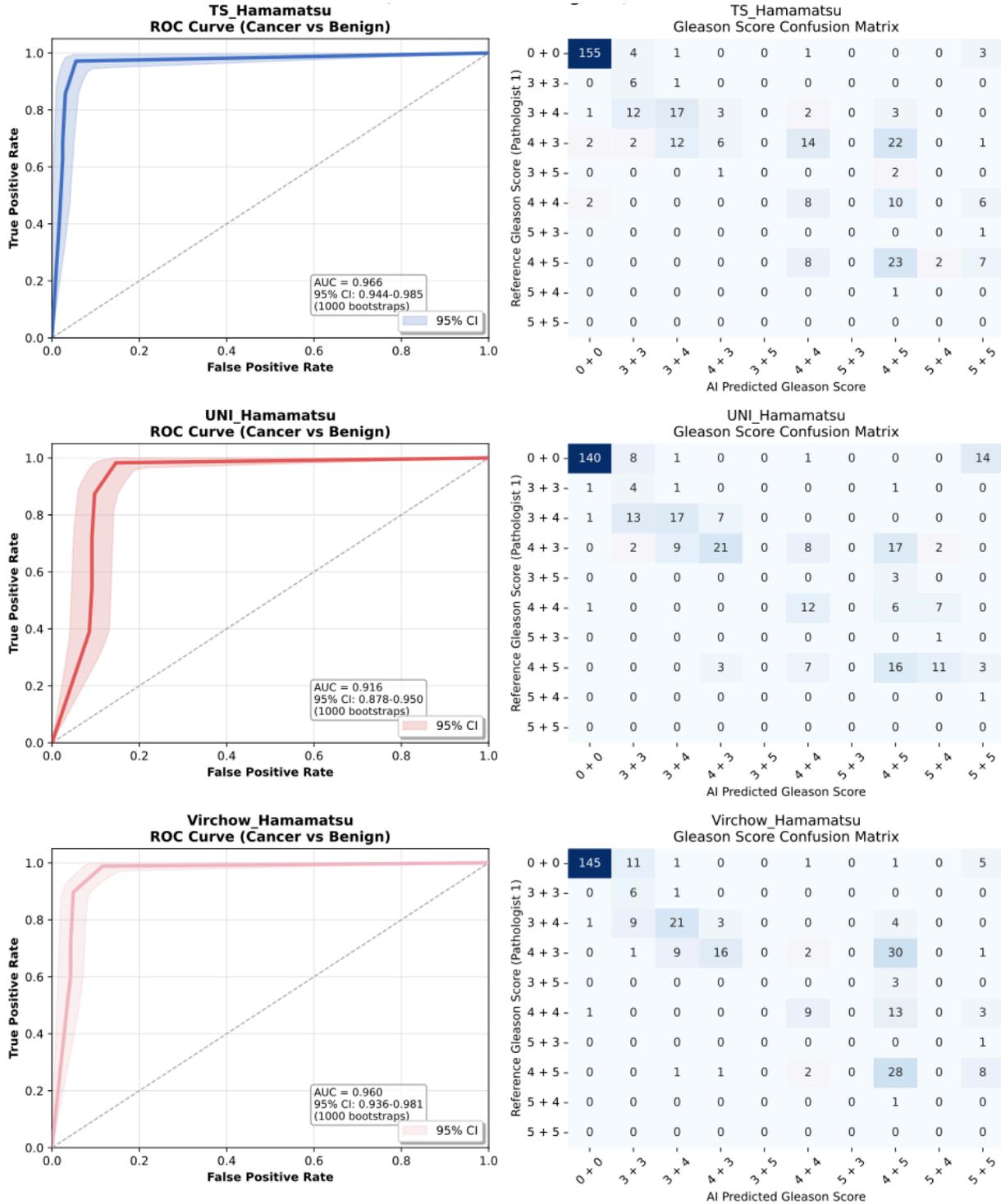

**Figure 2: Discriminatory performance and grading concordance of AI models (primary reference standard).** (Left) ROC analysis of the discriminatory performance between benign and malignant samples. Shaded regions represent 95% bootstrapped confidence intervals ($n$=337 slides). (Right) Confusion matrices comparing AI-predicted Gleason scores against the pathologist's grading. (Top to Bottom) AI models based on TS, UNI and Virchow2. All models were evaluated on images acquired with the Hamamatsu scanner and compared to the primary reference standard (pathologist I). AUC=area under the ROC curve, ROC=receiver operating characteristic, TS=task-specific model.

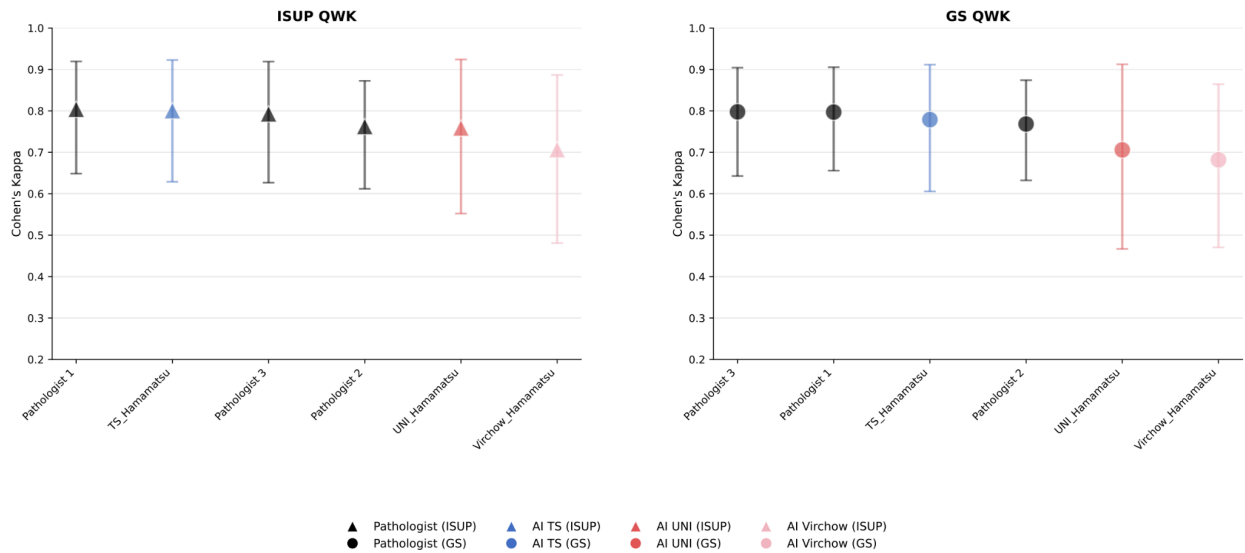

**Figure 3: Inter-observer agreement rankings (quadratically weighted kappa).** Mean pairwise Cohen's QWK for (Left) ISUP grade and (Right) Gleason score, evaluated on the slides assessed by all three pathologists (*n*=59). Markers represent the mean QWK of each observer (pathologist or AI model) calculated relative to the human pathologists. Error bars indicate 95% confidence intervals. Images acquired with the Hamamatsu scanner were used for the analysis. GS=Gleason score. ISUP=International Society for Urological Pathology. QWK=quadratically weighted kappa, TS=task-specific model.

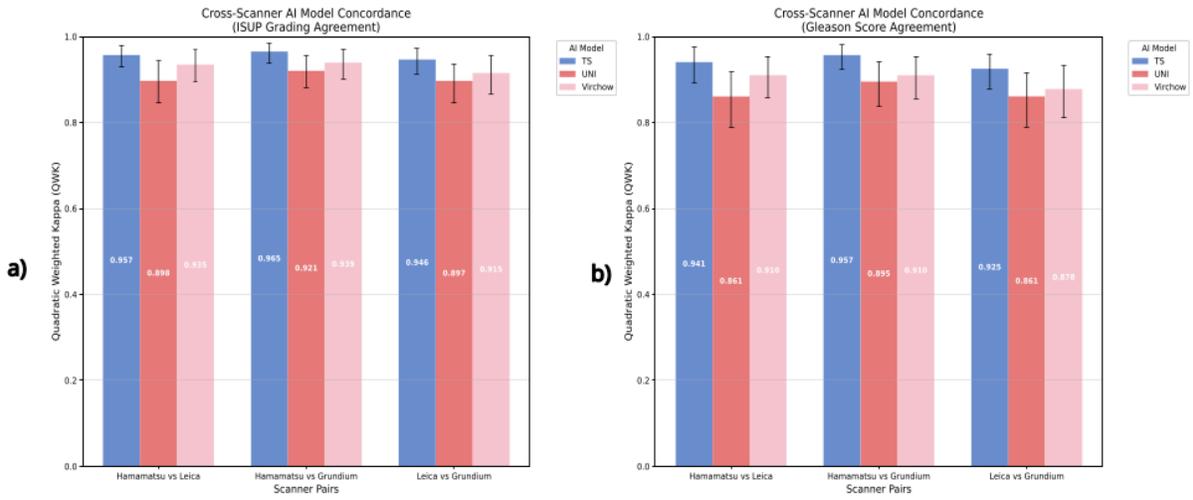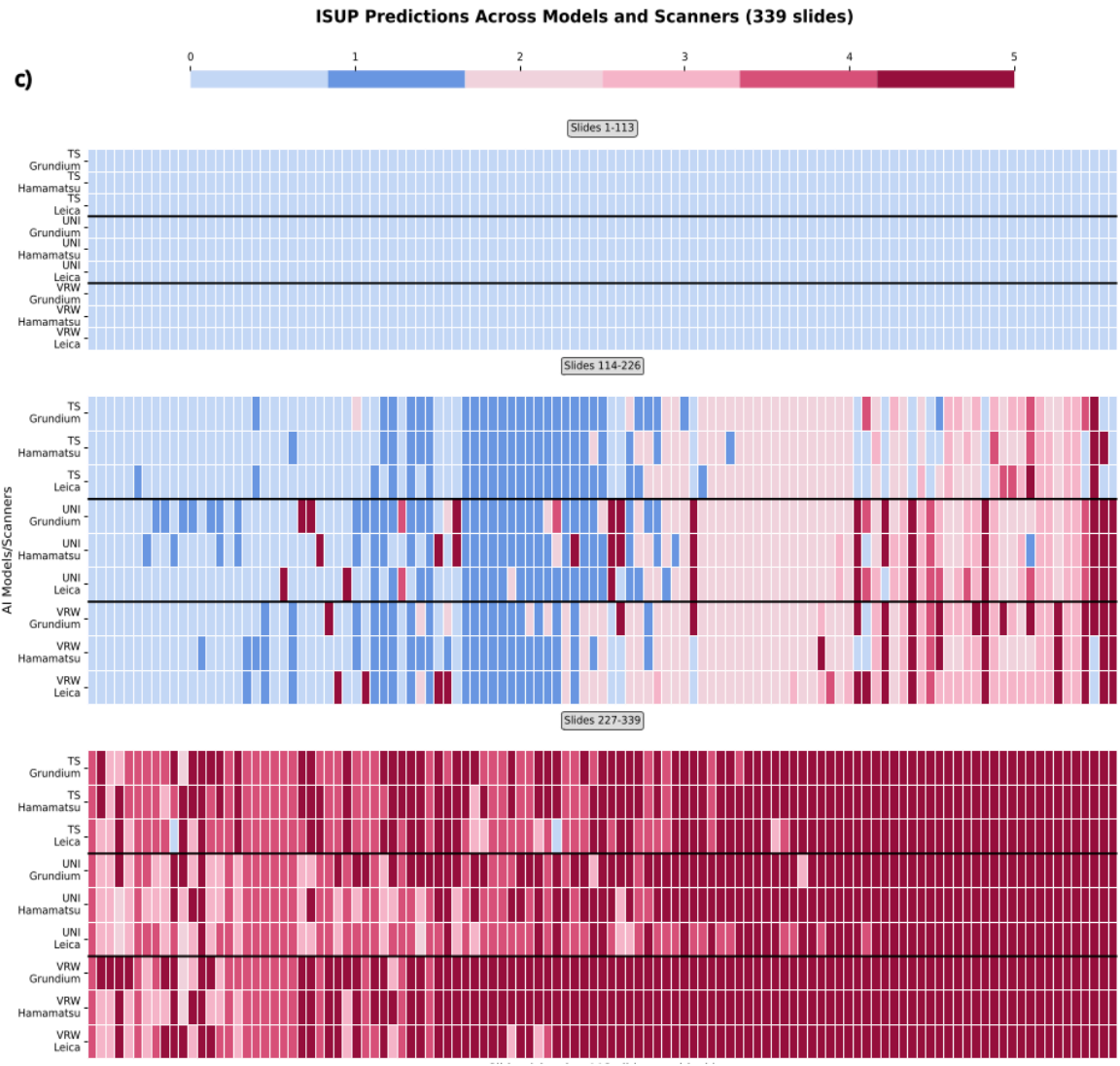

**Figure 4: Consistency of AI predictions of ISUP grade across scanner platforms.** (A, B) Cross-scanner concordance measured by Cohen's QWK for ISUP grade (A) and Gleason score (B). Bars represent concordance between scanner pairs (Hamamatsu vs Leica, Hamamatsu vs Grundium, and Leica vs Grundium) for each AI model. Error bars indicate 95% confidence intervals. (C) Heatmap of slide-level ISUP grade predictions across three scanners for all 339 slides. Columns represent individual slides; rows represent AI model/scanner combinations. Colors correspond to ISUP grades 0–5. ISUP=International Society for Urological Pathology, QWK=quadratically weighted kappa, TS=task-specific model, VRW=Virchow2 model.

# Validation of Diagnostic Artificial Intelligence Models for Prostate Pathology in a Middle Eastern Cohort

**Supplementary Appendix**

Peshawa J. Muhammad Ali[1,2*], Navin Vincent[3*], Saman S. Abdulla[4,5], Han N. Mohammed Fadhl[6], Anders Blilie[7,8], Kelvin Szolnoky[9], Julia Anna Mielcarz[3], Xiaoyi Ji[9], Nita Mulliqi[3], Abdulbasit K. Al-Talabani[1‡], Kimmo Kartasalo[3‡]

10. Department of Software Engineering, Faculty of Engineering, Koya University, Koya 44023, Kurdistan Region - F.R. Iraq
11. Department of Mechanical and Manufacturing Engineering, Faculty of Engineering, Koya University, Koya 44023, Kurdistan Region - F.R. Iraq
12. Department of Medical Epidemiology and Biostatistics, SciLifeLab, Karolinska Institutet, Stockholm, Sweden
13. College of Dentistry, Hawler Medical University, Erbil, Kurdistan Region, Iraq
14. PAR Private Hospital, Erbil, Kurdistan Region, Iraq
15. College of Dentistry, University of Sulaimani, Sulaymaniyah, Kurdistan Region, Iraq
16. Department of Pathology, Stavanger University Hospital, Stavanger, Norway
17. Faculty of Health Sciences, University of Stavanger, Stavanger, Norway
18. Department of Medical Epidemiology and Biostatistics, Karolinska Institutet, Stockholm, Sweden

* These authors contributed equally to this work.

‡ These authors jointly supervised the work.

Corresponding author: Kimmo Kartasalo, kimmo.kartasalo@ki.se

**Table of Contents**

## Supplementary Methods

**Tissue detection and tiling**

Tissue detection from WSIs was performed using a custom-built tissue segmentation model[1] based on a UNet++ architecture, incorporating a ResNet18 encoder.[2] Initially, 512×512px patches were extracted across the entire WSI at 8.0 μm/px resolution, with a 128 px overlap, followed by pixel-wise segmentation to identify tissue regions. These segmented regions were then combined into a single binary-tissue mask per WSI. Next, 256×256 px high-resolution tissue patches were extracted at 1.0 μm/px resolution, using the segmentation masks to retain only those patches where at least 10% of pixels contained tissue. During model training, patches were extracted without overlap to reduce GPU memory usage, whereas, for model prediction, a 128 px overlap was used to enhance diagnostic accuracy. To achieve a resolution of 1.0 μm/px, patches were downsampled from the nearest higher resolution level in the WSI resolution pyramid using Lanczos resampling. Each tile was stored as a compressed RGB image within an LMDB (Lightning Memory-Mapped Database) environment for efficient random access. To ensure modular storage and parallel data loading, each WSI was assigned its own LMDB file. Associated spatial and metadata information were stored in Parquet tables.

**Artificial intelligence model**

All evaluated AI models employ an attention-based multiple instance learning (ABMIL)[3] framework that uses only weakly supervised learning, leveraging slide-level labels. In the task-specific (TS) model,[4] patch-level features are extracted with an EfficientNet-V2-S[5] encoder and aggregated into slide-level representations with the ABMIL. Downstream linear classification layers are used to predict the two Gleason patterns (i.e. 3, 4, or 5), which are further translated into GS and ISUP grade. Training was conducted end-to-end, jointly optimising all parameters with cross-entropy loss and the AdamW optimiser[6] with a base learning rate of 0.0001. Full architectural choices, hyperparameters, and validation details are described in the original publication. The same pipeline was also run with the UNI and Virchow2 FMs,[7,8] keeping their encoders frozen and training only the ABMIL and terminal classification layers identically to the TS setup. We used 10-fold cross-validation stratified by patient and ISUP grade. In inference, test-time augmentation (TTA) was applied for each fold and model. For each slide, the model network predicts primary and secondary Gleason patterns and aggregates them to slide-level GS based on the output softmax vector. We map $GP \in \{0,1,2,3\}$ to Gleason score using a fixed rule (e.g. 1+1→3+3, 1+2→3+4, 2+1→4+3) and then to ISUP Grade Group (0–5).[9] The pre-specified diagnostic decision rule declares test positivity (cancer present) when ISUP ≥ 1 and test negativity when ISUP = 0. The final slide-level GS and ISUP grade output was the majority vote across the 10 models, with cancer probability taken as the ensemble median. Mean attention scores from the ABMIL models were used to highlight tile-level regions of interest informing the prediction. All AI inferences were performed solely on digital slide images without access to patient clinical information or reference standard results.

**Hardware and software**

We used Python 3.8.10 with PyTorch 2.0.0 (CUDA 12.2)[10] and PyTorch DDP[11] for multi-GPU training across all experiments and models. UNI[7] and Virchow2[8] foundation models were initialised from their HuggingFace releases and paired with timm 0.9.8 ViT implementations,[12] keeping encoder weights as provided. Containerisation used Docker 20.10.21 locally and Singularity/Apptainer on compute clusters.[13,14] WSIs were accessed via OpenSlide 4.0.0,[15] openslide-python 1.3.1, and OpenPhi 2.1.0.[16] Image augmentations employed Albumentations 1.3.1 and Stainlib 0.6.1.[17] The tissue segmentation model was implemented with segmentation_models_pytorch 0.3.3.[18] The LMDB data access framework was implemented in Python using the lmdb library (1.5.1) for memory-mapped storage of compressed RGB tiles and Polars 1.8.2 for high-performance handling of Parquet metadata tables.[19] Numerical computing, evaluation, and data handling used NumPy 1.24.0,[20] scikit-learn 1.2.2,[21] and Pandas 1.5.3;[22] basic image processing used Pillow 9.4.0 and OpenCV-python. Visualisations were produced with Matplotlib 3.7.1 and Seaborn 0.12.2,[23,24] and figure panels assembled in InkScape.

# Supplementary Figures

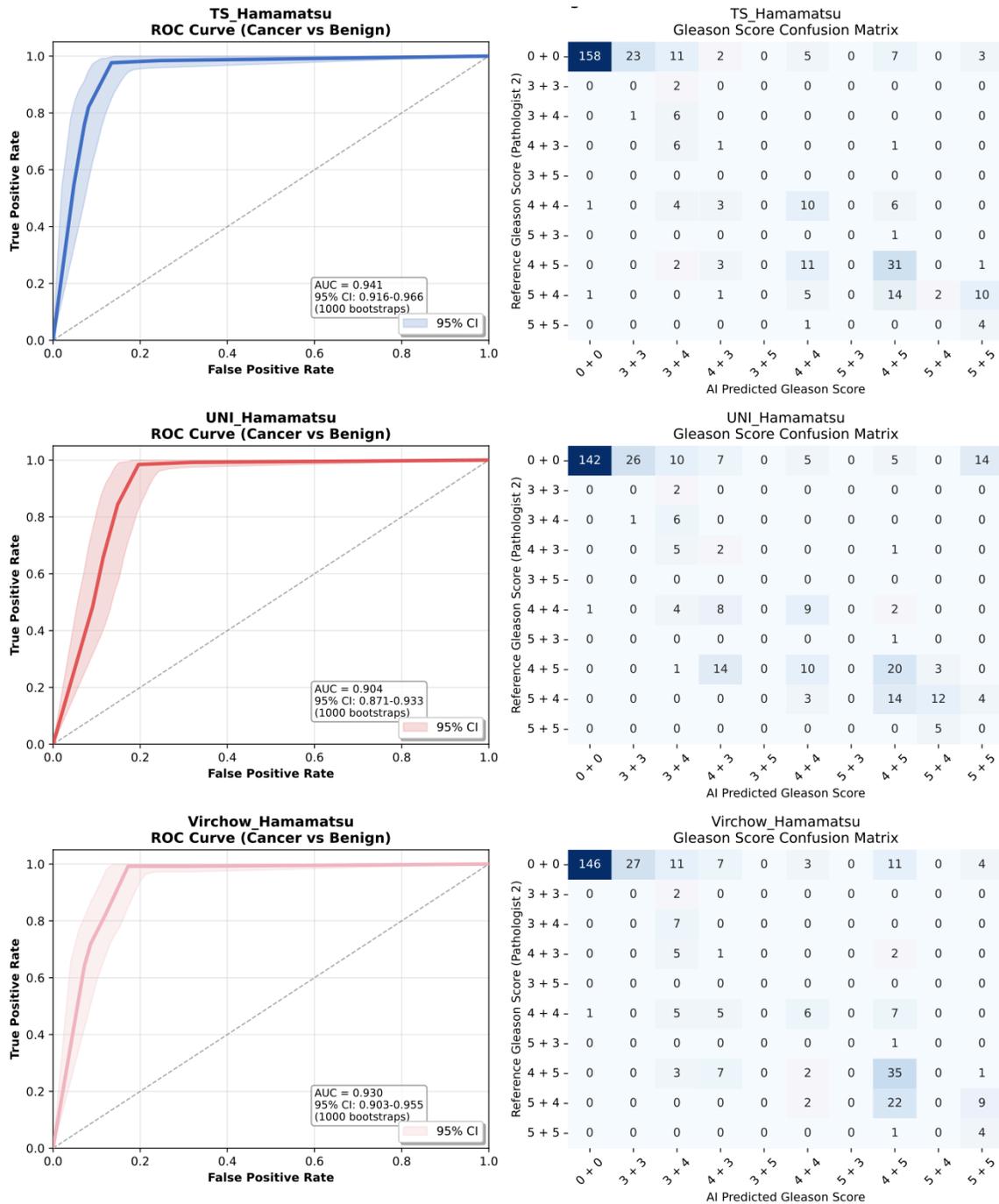

**Supplementary Figure 1: Discriminatory performance and grading concordance of AI models (secondary reference standard).** (Left) ROC analysis of the discriminatory performance between benign and malignant samples. Shaded regions represent 95% bootstrapped confidence intervals ($n$=337 slides). (Right) Confusion matrices comparing AI-predicted Gleason scores against the pathologist's grading. (Top to Bottom) AI models based on TS, UNI and Virchow2. All models were evaluated on images acquired with the Hamamatsu scanner and compared to the secondary reference standard (pathologist II). AUC=area under the ROC curve, ROC=receiver operating characteristic, TS=task-specific model.

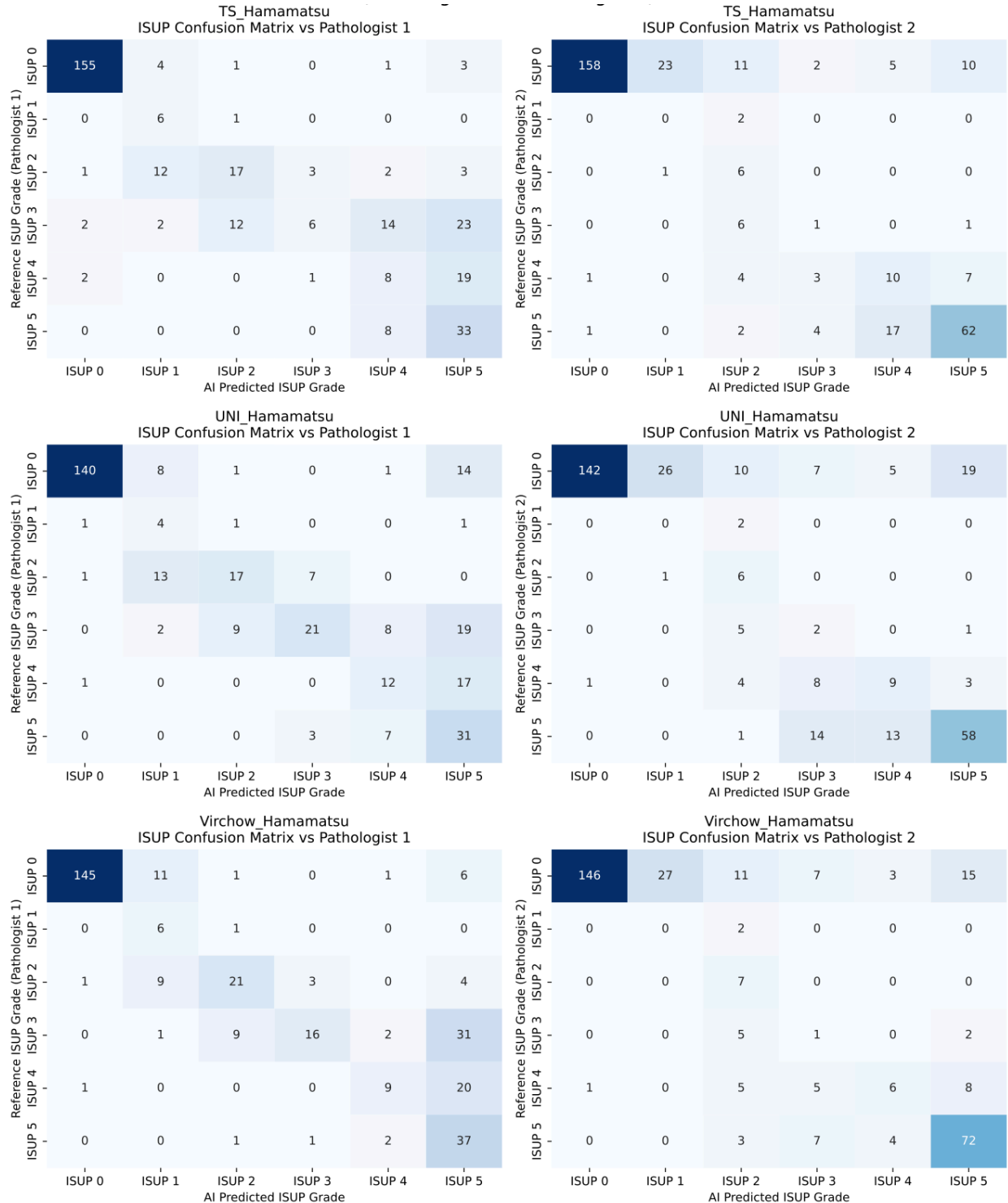

**Supplementary Figure 2: Confusion matrices for ISUP grade predictions.** Comparisons of AI model predictions of ISUP grade against primary reference standard, pathologist I (Left) and secondary reference standard, pathologist II (Right). (Top to Bottom) AI models based on TS, UNI and Virchow2. All models were evaluated on images acquired with the Hamamatsu scanner. ISUP=International Society for Urological Pathology, TS=task-specific model.

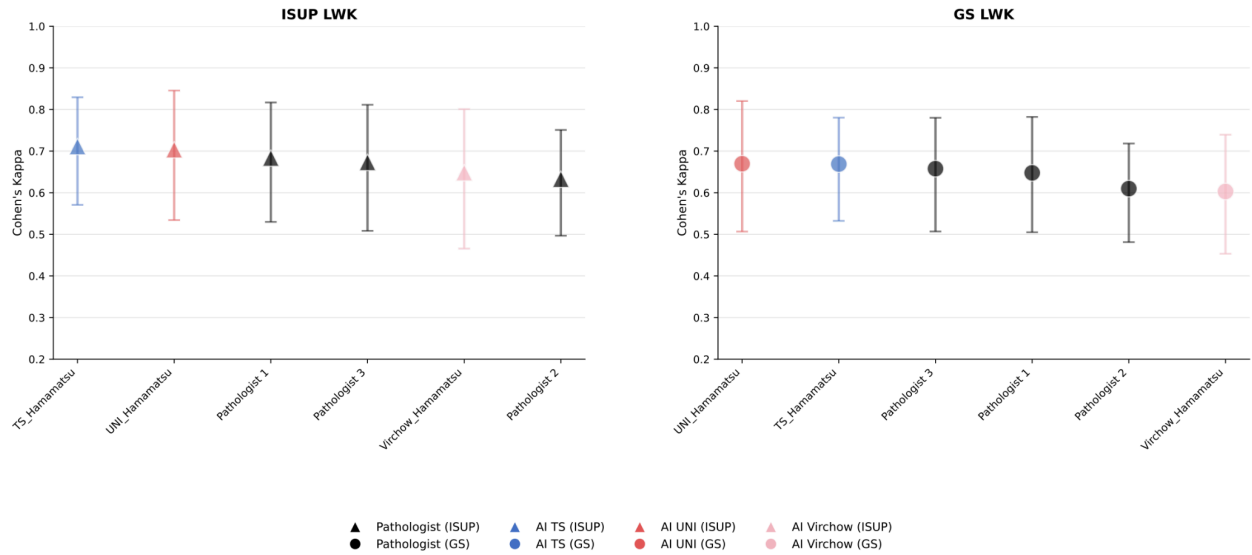

**Supplementary Figure 3: Inter-observer agreement rankings (linearly weighted kappa).** Mean pairwise Cohen's LWK for (Left) ISUP grade and (Right) Gleason score, evaluated on the slides assessed by all three pathologists (*n*=59). Markers represent the mean LWK of each observer (pathologist or AI model) calculated relative to the human pathologists. Error bars indicate 95% confidence intervals. Images acquired with the Hamamatsu scanner were used for the analysis. GS=Gleason score. ISUP=International Society for Urological Pathology. LWK=linearly weighted kappa, TS=task-specific model.

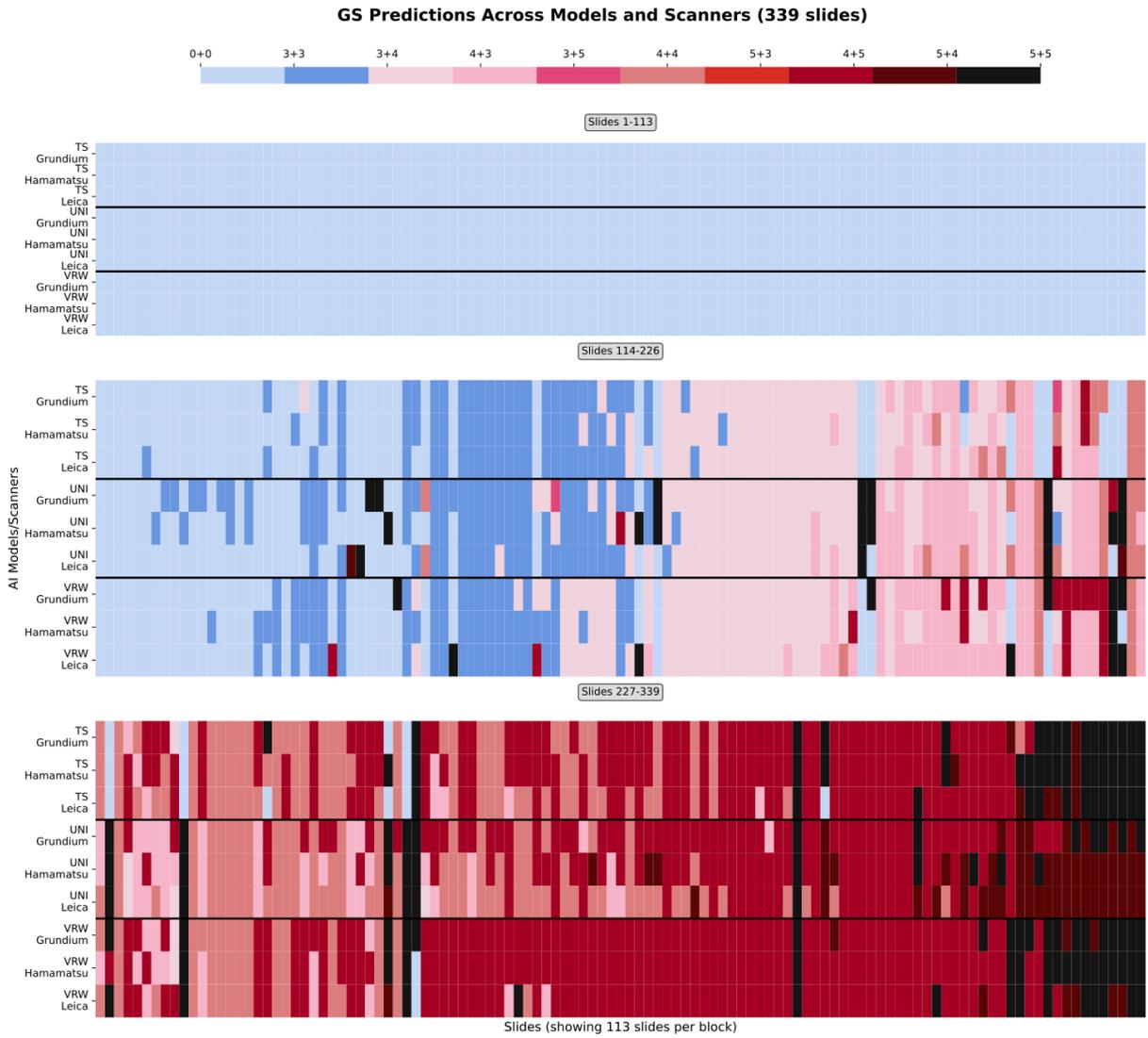

**Supplementary Figure 4: Consistency of AI predictions of Gleason score across scanner platforms.** Heatmap of slide-level GS predictions across three scanners for all 339 slides. Columns represent individual slides; rows represent AI model/scanner combinations. Colors correspond to Gleason scores. GS=Gleason score, TS=task-specific model, VRW=Virchow2 model.

## Supplementary Tables

**Supplementary Table 1: Overview of external validation studies for commercial or research-grade AI prostate pathology systems.** The table lists the source region and size of the cohort, the software/model, and the scanner hardware. WSI=whole-slide image.

| Validation study | Software or AI model | Cohort source | Scanner model | Cohort size |
|---|---|---|---|---|
| Raciti et al., 2020[25] | Paige Prostate | USA | Leica Aperio AT2 | 232 patients, 232 slides |
| Perinchiri et al. 2021[26] | Paige Prostate | USA | Leica Aperio AT2 | 118 patients, 1,876 cores |
| Da-silva et al, 2021[27] | Paige Prostate | Brazil | Leica Aperio AT2 | 100 patients |
| Jung et al., 2022[28] | DeepBio Dx | South Korea | Leica Aperio AT2 | 593 patients |
| Tolkach et al., 2023[29] | Halo Prostate | Germany & Austria | Leica Aperio GT450, Hamamatsu NanoZoomer S360, Hamamatsu NanoZoomer C9600-12 | The number of patients is not clear: 5,847 slides |
| Vazzano et al., 2023[30] | INIFY Prostate | USA | Leica Aperio ImageScope, Hamamatsu, Philips | 30 patients, 90 slides |
| Santa-Rosario et al., 2024[31] | IBEX Galen Prostate | USA | 3DHISTECH PANNORAMIC P-250 | 101 patients |
| Lami et al., 2024[32] | IBEX Galen Prostate | Japan | Philips Ultrafast Scanner | 100 cases (697 WSIs) |

**Supplementary Table 2: Age-standardised incidence, mortality, and mortality-to-incidence ratio (MIR) for prostate cancer in selected countries and regions.** Values are based on international cancer registry estimates (Global Cancer Observatory). ASR=age-standardised rate per 100,000 population.

| Countries/Regions | ASR Incidence | ASR Mortality | MIR |
|---|---|---|---|
| USA | 75.2 | 8.1 | 0.11 |
| Sweden | 104.3 | 13.4 | 0.13 |
| Northern Europe | 82.8 | 12.4 | 0.15 |
| The Netherlands | 57 | 11.4 | 0.20 |
| Western Asia | 24 | 7.9 | 0.33 |
| Iraq | 12.7 | 4.9 | 0.39 |

**Supplementary Table 3: Cancer detection and grading performance metrics of AI models (QWK, secondary reference standard).** Specificity, sensitivity, and Cohen's kappa for ISUP grade and Gleason score for three AI systems (task-specific model, UNI foundation model, Virchow2 foundation model). Concordance was computed relative to pathologist II who set the secondary reference standard. QWK values are reported for all slides (overall, $n=337$) and for malignant slides ($n=128$) only. Confidence intervals represent 95% bootstrapped estimates. GS=Gleason score, ISUP=International Society of Urological Pathology, QWK=quadratically weighted kappa.

| AI model | Specificity | Sensitivity | QWK-ISUP (Overall) | QWK-ISUP (Malignant) | QWK-Gleason (Overall) | QWK-Gleason (Malignant) |
|---|---|---|---|---|---|---|
| Task-specific | 0.756 (0.695-0.813) | 0.984 (0.959-1.000) | 0.831 (0.771-0.883) | 0.602 (0.443-0.752) | 0.822 (0.752-0.879) | 0.690 (0.566-0.779) |
| UNI | 0.679 (0.618-0.742) | 0.992 (0.975-1.000) | 0.744 (0.670-0.813) | 0.594 (0.468-0.708) | 0.702 (0.617-0.786) | 0.644 (0.523-0.734) |
| Virchow2 | 0.699 (0.634-0.757) | 0.992 (0.972-1.000) | 0.793 (0.731-0.846) | 0.616 (0.471-0.742) | 0.792 (0.727-0.850) | 0.672 (0.549-0.770) |

**Supplementary Table 4: Grading performance metrics of AI models (LWK, primary reference standard).** Specificity, sensitivity, and Cohen's kappa for ISUP grade and Gleason score for three AI systems (task-specific model, UNI foundation model, Virchow2 foundation model). Concordance was computed relative to pathologist I who set the primary reference standard. LWK values are reported for all slides (overall, $n=339$) and for malignant slides ($n=175$) only. Confidence intervals represent 95% bootstrapped estimates. GS=Gleason score, ISUP=International Society of Urological Pathology, LWK=linearly weighted kappa.

| AI model | LWK-ISUP (Overall) | LWK-ISUP (Malignant) | LWK-GS (Overall) | LWK-GS (Malignant) |
|---|---|---|---|---|
| TS | 0.764 (0.720-0.804) | 0.489 (0.405-0.563) | 0.687 (0.637-0.729) | 0.445 (0.371-0.523) |
| UNI | 0.723 (0.670-0.774) | 0.522 (0.445-0.595) | 0.641 (0.581-0.702) | 0.477 (0.404-0.554) |
| Virchow2 | 0.760 (0.713-0.800) | 0.489 (0.399-0.572) | 0.681 (0.620-0.731) | 0.413 (0.336-0.493) |

**Supplementary Table 5: Grading performance metrics of AI models (LWK, secondary reference standard).** Specificity, sensitivity, and Cohen's kappa for ISUP grade and Gleason score for three AI systems (task-specific model, UNI foundation model, Virchow2 foundation model). Concordance was computed relative to pathologist II who set the secondary reference standard. LWK values are reported for all slides (overall, $n=337$) and for malignant slides ($n=128$) only. Confidence intervals represent 95% bootstrapped estimates. GS=Gleason score, ISUP=International Society of Urological Pathology, LWK=linearly weighted kappa.

| AI model | LWK-ISUP (Overall) | LWK-ISUP (Malignant) | LWK-Gleason (Overall) | LWK-Gleason (Malignant) |
|---|---|---|---|---|
| TS | 0.751 (0.700-0.805) | 0.498 (0.377-0.688) | 0.728 (0.677-0.777) | 0.524 (0.423-0.617) |
| UNI | 0.666 (0.604-0.720) | 0.469 (0.354-0.580) | 0.632 (0.568-0.694) | 0.480 (0.384-0.576) |
| Virchow2 | 0.715 (0.659-0.772) | 0.515 (0.386-0.635) | 0.695 (0.638-0.752) | 0.507 (0.396-0.608) |

**Supplementary Table 6: Distribution of alternative (non-malignant) clinical diagnoses among patients with ISUP < 2.** Frequency counts of clinician-recorded alternative diagnoses (for patients whose final ISUP grade was < 2) as recorded in pathology/clinical notes. This table summarises the set of common non-malignant diagnostic labels present in the cohort and their approximate frequencies. ISUP=International Society for Urological Pathology, LUTS=lower urinary tract symptoms.

| Alternative diagnosis | Frequency |
|---|---|
| Benign prostatic hyperplasia | ~6–7 |
| Prostatic nodules | ~5–6 |
| Hard / firm prostate | ~5 |
| LUTS / obstructive symptoms | ~2–3 |
| Suspicious area (no cancer) | ~2 |
| Other (plasmacytoma, etc.) | 1 |

**Supplementary Table 7: Representative examples of alternative diagnoses extracted from clinical notes.** Representative text excerpts and short free-text examples from pathology/clinical notes illustrating the kinds of alternative diagnoses recorded for patients with ISUP grade < 2. BPH=benign prostatic hyperplasia, ISUP=International Society for Urological Pathology, LUTS=lower urinary tract symptoms, TRUS=transrectal ultrasound, TURP=transurethral resection of the prostate, US=ultrasound.

| Category | Representative examples (from "Notes") |
|---|---|
| Benign prostatic hyperplasia | "BPH exists", "Mild BPH with nodules", "BPH 125 gm", "Enlarged prostate", "TURP in 2010" |
| Prostatic nodules | "Left prostatic nodule", "Right side hard nodule", "Prostatic nodules on US (over 15 nodules)" |
| LUTS / Obstructive symptoms | "LUTS with hard nodule in the prostate", "Obstructive urinary" |
| Suspicious but non-malignant findings | "Highly suspicious irregular hypoechoic bulge", "Suspicious area in the right posterolateral prostate" |
| Firm or stony hard glands | "Stony hard prostate", "Firm right prostate gland", "Hard prostate, TRUS" |
| Miscellaneous / rare | "Plasmacytoma (left prostatic mass)", "Pinal masses" |

**Supplementary Table 8: Review of slides with clinically significant AI errors.** Slide-level summary of cases where any AI model exhibited clinically significant errors relative to both pathologist I and pathologist II, in accordance with the pre-specified error definition (benign predicted as ISUP ≥ 2, or ISUP ≥ 2 predicted as benign). For each slide the table lists IDs, Gleason scores assigned by pathologists I and II and by each AI model (TS, UNI, Virchow2). For slides where at least two out of three models exhibited a significant error, a third Gleason scoring by pathologist III and a free-text assessment is shown. GS=Gleason score, TS=task-specific model, UFM = UNI foundation model, VFM = Virchow2 foundation model.

| ID | GS Path. I | GS Path. II | GS TS | GS UFM | GS VFM | GS Path. III | Comment (Path. 3) |
|---|---|---|---|---|---|---|---|
| C017A | 0+0 | 0+0 | 0+0 | 5+5 | 5+5 | 5+4 | Thick cut and markedly overstained tissue. Not possible to evaluate nuclear/cytological features - assessment primarily based on architecture. Very hard to evaluate in many areas - there is some obvious inflammation; presence of acute inflammatory cells distinctly different from prostatic epithelium. But also cells too large and atypical to represent granulocytes or lymphocytes - favors cancer. Single cells, solid areas, and poorly formed glands. Would do immunostaining (PSA/PSMA/PANCK) to confirm and assess the extent of cancer. If possible: new slide preparation with better quality. |
| C017B | 0+0 | 0+0 | 0+0 | 5+5 | 0+0 | - | Single model error (UNI). |
| C018B | 0+0 | 0+0 | 0+0 | 5+5 | 0+0 | - | Single model error (UNI). |
| C029A | 0+0 | 0+0 | 0+0 | 5+5 | 0+0 | - | Single model error (UNI). |
| C033A | 0+0 | 0+0 | 0+0 | 5+5 | 0+0 | - | Single model error (UNI). |
| C043A | 4+4 | 4+4 | 0+0 | 0+0 | 0+0 | 4+5 | Thick cut and markedly overstained tissue. Not possible to evaluate nuclear/cytological features - assessment primarily based on architecture. Areas of tightly packed small, dark cells interpreted as inflammation/lymphoid aggregates (although one could argue for cancer due to gland-like openings/slit spaces here). Within the eosinophilic stroma there are numerous cells with pale cytoplasm, interpreted as cancer. Partly poorly formed glandular structures, partly single cells. Also areas of solid, seemingly intraductal proliferations and one gland with cribriform appearance, but they do not necessarily look infiltrative (IDC possible). I believe the pale areas in the stroma represent true cancer, but it is difficult to rule out inflammation with some sort of artifact producing the paleness in these parts. Immunostaining to confirm epithelial origin of the pale-area cells would be necessary (e.g. PANCK or PSA). |
| C066A | 0+0 | 0+0 | 5+5 | 5+5 | 0+0 | 5+5 | Slide not as overstained as others, but still hard to grasp nuclear features which makes assessment difficult. There is definitive inflammation, but also a high number of stromal single cells that are more atypical than acceptable for lymphocytes or fibroblasts. Would do IHC (PANCK/PSA). |
| C066B | 0+0 | 0+0 | 5+5 | 5+5 | 5+5 | 5+5 | Slide not as overstained as others, but still hard to grasp nuclear features which makes assessment difficult. There is a marked increase of cells in the stroma which are difficult to place in the inflammatory vs. cancer category. Definitively some of it represents inflammation, including many seemingly eosinophilic cells, but also large atypical cells which I believe to be malignant. Would do IHC (PANCK/PSA). |
| C070 | 0+0 | 0+0 | 4+4 | 4+4 | 4+4 | 4+5 | Slide not as overstained as others, but still hard to grasp nuclear features which makes assessment difficult. It is easy to see that there is *some* definitive cancer in this slide - certain areas of solid/cribriform tumor. But pale areas and single stromal cells are hard to assess in terms of inflammation vs. cancer. There are some large atypical cells within these areas, favoring Gleason pattern 5 cancer. Would do IHC (PANCK/PSA). |
| C084 | 0+0 | 0+0 | 0+0 | 5+5 | 4+5 | 4+5 | Thick cut and overstained tissue. Not possible to evaluate nuclear/cytological features - assessment primarily based on architecture. Stromal cells are hard to assess in terms of inflammation vs. cancer. Some areas of seemingly poorly |
| | | | | | | | formed glands of Gleason pattern 4, plus single cells that seem too large and atypical for inflammation. Would do IHC (PANCK/PSA). |
| C089A | 0+0 | 0+0 | 0+0 | 5+5 | 0+0 | - | Single model error (UNI). |
| C089B | 0+0 | 0+0 | 5+5 | 5+5 | 5+5 | 5+4 | Thick cut and overstained tissue. Not possible to evaluate nuclear/cytological features - assessment primarily based on architecture. Marked increase of cells in the stroma which are difficult to place in the inflammatory vs. cancer category. Definitively some of it is inflammation, including many macrophages and possible granulomas, but also large atypical cells which I believe to be malignant. Would do IHC (PANCK/PSA). |
| C140 | 0+0 | 0+0 | 0+0 | 5+5 | 0+0 | - | Single model error (UNI). |
| C160A | 0+0 | 0+0 | 0+0 | 5+5 | 0+0 | - | Single model error (UNI). |
| C165A | 0+0 | 0+0 | 3+3 | 3+4 | 3+3 | 3+3 | Slide not as overstained as others, but still hard to grasp nuclear features which makes assessment difficult. There are some areas of low-grade 3+3 cancer. No definitive high-grade cancer, but it is hard to assess areas of increased cellularity within the stroma (inflammation vs. malignancy). Could just be inflammation - not enough atypia in these cells to justify a diagnosis of high-grade cancer. There are also pale areas which are hard to assess - leaning towards benign in these areas. There is an area of solid epithelium within glands, but this looks more like squamous metaplasia than cancer. Would do IHC (PANCK/PSA). |
| C170B | 0+0 | 0+0 | 0+0 | 5+5 | 5+5 | 4+5 | Slide not as overstained as others, but still hard to grasp nuclear features which makes assessment difficult. Much of the tissue is normal, without cancer or inflammation. There is an area of Gleason pattern 4 cancer that is possible to assess. The main problem is the areas of increased single cells in the stroma, where inflammation vs. G5 cancer is more or less a coin-toss decision. Favoring cancer due to some atypia, but would need confirmatory IHC (PANCK/PSA). |

**Supplementary Table 9: Adherence to STARD-AI reporting items for diagnostic accuracy studies of AI.**
Item-by-item STARD-AI[33] checklist mapping: each reporting item is linked to the manuscript page/section where it is addressed and judged as "Satisfied", "Partially satisfied", or "Not applicable". This table documents compliance with reporting standards for AI diagnostic accuracy studies and highlights areas where reporting could be strengthened for transparency and reproducibility.

| No. | STARD-AI Item | Location (Page/ Section) | Context/ Evidence | Status |
|---|---|---|---|---|
| 1 | Identification as AI diagnostic accuracy study in title/abstract | p. 1 Title; p. 2 Abstract | Title includes "Diagnostic Artificial Intelligence Models"; abstract describes diagnostic accuracy metrics (AUC, sensitivity, specificity). | Satisfied |
| 2 | Structured abstract with design, methods, results, conclusions | p. 2 Abstract | Headings ("Background / Methods / Findings / Interpretation / Funding"). Clear design & metrics. | Satisfied |
| 3 | Scientific & clinical background; intended use of index test | pp. 4-5 Introduction | Discusses exclusion of under-represented groups, and explains digital pathology context, inequities, and the purpose of AI models (TS, UNI, Virchow) for prostate biopsy diagnosis. | Satisfied |
| 4 | Study objectives & hypotheses | p. 5 end of Introduction | "We hypothesised that (i) the AI systems would perform comparably... (ii) their predictions would remain consistent...". | Satisfied |
| 5 | Prospective / retrospective design | p. 6 Methods - Study design and participants | "The dataset represents a consecutive series …" and "... specimens in a retrospective setting." | Satisfied |
| 6 | Ethics approval / waiver | p. 6 Methods - Study design and participants | "PAR Hospital (permit 1002/07072024)... Swedish Ethical Review Authority (permit 2019-05220)" | Satisfied |
| 7 | Inclusion / exclusion criteria | p. 6 Methods - Study design and participants | Describes restriction to needle biopsies and exclusion of prostatectomy/TURP. | Satisfied |
| 8 | Basis for identifying eligible participants | p. 6 Methods - Study design and participants | "All pathology reports mentioning prostate cancer were retrieved... restriction to suspected adenocarcinoma … core biopsies..." | Satisfied |
| 9 | Setting/location / dates | p. 6 Methods - Study design and participants | "PAR Private Hospital, Erbil, Kurdistan Region of Iraq... between mid-2013 to mid-2024." | Satisfied |
| 10 | Consecutive / random / convenience series | p. 6 Methods - Study design and participants | "Consecutive series... included without any sampling". | Satisfied |
| 11 | Source of data and collection purpose | p. 6 Methods - Study design and participants | Retrospective routine archival data collected for diagnostic use then repurposed for validation. | Satisfied |

| 12 | Who annotated / experience | p. 6-7 Methods - Reference standard | Three experienced pathologists (two Iraq, one Europe) grading blinded slides in Cytomine. | Satisfied |
|---|---|---|---|---|
| 13 | Devices and software | p. 6 Methods - Slide scanning; p. 7 Methods - Artificial intelligence model | Lists Grundium Ocus, Hamamatsu HT 2.0, Leica Aperio; TS, UNI, Virchow2. | Satisfied |
| 14 | Data acquisition & pre-processing protocols | p. 6 Methods - Slide scanning; p. 7 Methods - Artificial intelligence model | "Scanned at 40× magnification, yielding resolutions of 0.22... 0.26... 0.25 µm/pixel" UNet++ segmentation, patch tiling 256x256. | Satisfied |
| 15a | Index test description for replication | p. 7 Methods - Artificial intelligence model | Describes ABMIL architecture, EfficientNet-V2-S encoder, loss functions, training scheme. | Satisfied |
| 15b | Development & training / validation sets & sizes | p. 7 Methods - Artificial intelligence model; p. 2 Abstract - Methods | "In the TS model, patch-level … The same design was applied to the UNI and Virchow2 FMs keeping their encoders frozen and training the ABMIL and classification layers identically to the TS setup." These models are all cited properly. For the external validation set "We collected and digitised 339 prostate biopsy specimens." | Satisfied |
| 15c | Definition of cut-offs for positivity | p. 3 Supplementary Methods - Artificial intelligence model | "Test positivity (cancer present) when ISUP ≥ 1; negativity when ISUP = 0". | Satisfied |
| 15d | End-user and required expertise | p. 4 Introduction | Models serve as decision support; implied in "support health care struggling with pathology workforce shortages". | Satisfied |
| 16a | Reference standard detail for replication | p. 6-7 Methods - Reference standard | Three pathologists, blinded ISUP/Gleason grading on Hamamatsu WSIs via Cytomine. | Satisfied |
| 16b | Rationale for reference standard | p. 4 (Introduction) | States "Gleason grading of biopsies is crucial in therapeutic decision-making", implying it is the established clinical standard. | Satisfied |
| 16c | Cut-offs in reference standard | p. 7 (Methods - Reference standard) | Uses ISUP grading system (0-5) based on Gleason scores. | Satisfied |
| 17a | Whether performers of index test had access to clinical info / reference results | p. 7 (Methods - Artificial intelligence model) | "All AI inferences performed solely on digital images without access to patient info or reference results." | Satisfied |
| 17b | Whether assessors of reference standard had access to index test results | p. 7 (Methods - Reference standard) | "Pathologists were blinded to the patients' clinical characteristics... and AI outputs". | Satisfied |
| 18 | Methods for estimating diagnostic accuracy | p. 8 (Methods - Statistical analysis) | "Sensitivity... specificity... AUC... Cohen's quadratically weighted kappa" with bootstrapped CIs. | Satisfied |
| 19 | Handling of indeterminate results | p. 11 ( Methods / Statistical analysis) p. 9 (Methods / Reference standard) | Text states "No indeterminate or AI-ungradable slides were encountered." Text states "S.A. graded all 339 slides, H.M. graded 337 slides..." | Satisfied |
| 20 | Handling of missing data | p. 16 Table 1 | States missing = 0 for slides | Satisfied |
| 21 | Analyses of variability (prespecified vs exploratory) | p. 8 (Methods- Statistical analysis) p. 10 (Results - Scanner consistency analysis) | Reports bootstrapped CI and scanner variability analysis. | Satisfied |
| 22 | Intended sample size and determination | p. 8 (Methods- Statistical analysis) | Sample size was dictated by data availability. | Satisfied |
| 23 | Algorithmic bias and fairness assessment | pp. 3,5 (Introduction) | Discusses equity motivation and underrepresented setting validation. The work aims to prove the generalisability of the models in the underrepresented populations. | Satisfied |
| 24 | Participant flow diagram | p. 18 (Figure 1) | Diagram from 502 cases -> 185 patients -> 339 slides. | Satisfied |
| 25 | Baseline demographic / technical characteristics | p. 16 (Table 1) | Age distribution, ISUP grades, slide counts, scanner types. | Satisfied |
| 26a | Disease severity distribution | p. 16 (Table 1) | ISUP grades 0-5 shown with percentages. | Satisfied |
| 26b | Alternative diagnoses distribution | p. 7 ( Methods - Reference standard) Supplementary Table 6-7 | "Alternative diagnoses were reported... (Supplementary Table 6-7)" | Satisfied |
| 27 | Interval and interventions between index and reference tests | p. 6 (Methods / Study design and participants) | The study is retrospective using archival material (2013–2024). AI and pathologists assessed the same digital images; no clinical interventions occurred between these assessments. | Satisfied |
| 28 | Dataset represents intended-use population | p. 6 (Methods / Study design and participants) | "Mirrors the intended-use population in this region". | Satisfied |
| 29 | Differences between external and training sets | p. 6 (Methods / Study design and participants) | Contrasts Iraqi single-centre vs multi-country Western training set. | Satisfied |
| 30 | Cross-tabulation of index vs reference results | p. 19 (Figure 2) Supplementary Figure 1 and Figure 2. | Confusion matrices comparing AI vs Pathologist. | Satisfied |
| 31 | Estimates of diagnostic accuracy + precision | pp. 8-10 (Results) p. 17 (Table 2) | Sensitivity/specificity with 95% CIs, QWK values shown. | Satisfied |

| 32 | Adverse events from tests | N/A | Retrospective/Non-invasive | N/A |
|---|---|---|---|---|
| 33 | Study limitations, bias, uncertainty, and generalisability | pp. 10-12 (Discussion) | Discusses observer variability, scanner differences, and generalisation limits. | Satisfied |
| 34 | Implications for practice / intended use | p. 10 Discussion opening | Emphasises AI use in resource-limited labs and equity implications. | Satisfied |
| 35 | Ethical issues of AI use / fairness | pp. 4-5 (Introduction); p. 10 (Discussion) | Addresses equity and bias context; ethical use implicitly covered. | Satisfied |
| 36 | Registration number and registry | p. 5 (end of Introduction); p. 15 Ref 27 | Refers to pre-registered BMJ Open study protocol (BMJ Open 2025 e097591) | Satisfied |
| 37 | Access to full protocol | p. 15 Ref 27 | Public protocol accessible via BMJ Open link. | Satisfied |
| 38 | Funding sources and roles | p. 2 (Funding); p. 12 (Acknowledgments) | Lists SciLifeLab, Wallenberg, KI Research Foundation; "funder had no role..." | Satisfied |
| 39 | Commercial interests | p. 12 (Declaration of interests) | "N.M. and K.K. are shareholders of Clinsight AB. Other authors declare no competing interests." | Satisfied |
| 40a | Availability of datasets and code | p. 7 Methods / Artificial intelligence model p. 12 Data sharing | Dataset in BioImage Archive, CC BY 4.0 license. No significant custom code developed for this study. | Satisfied |
| 40b | Auditability / stored outputs | p. 12 Data sharing | All AI predicted outputs are available as supplementary material. | Satisfied |

# Supplementary References